\documentclass[letterpaper, 10 pt, conference]{ieeeconf}  % Comment this line out if you need a4paper

\IEEEoverridecommandlockouts                              % This command is only needed if 
\usepackage{graphicx} % 
\usepackage{epsfig} % for postscript graphics files
\usepackage{mathptmx} % assumes new font selection scheme installed
\usepackage{mathpazo}
\usepackage{times} % assumes new font selection scheme installed
\usepackage{amsmath} % assumes amsmath package installed
\usepackage{amssymb}  % assumes amsmath package installed
\usepackage{algorithm}
\usepackage{algorithmic}
\usepackage{setspace}
\usepackage{pifont}
\usepackage{multirow}
\usepackage{multicol}

\title{\LARGE \bf
DuEqNet: Dual-Equivariance Network in Outdoor 3D Object Detection for Autonomous Driving
}

\author{
Xihao Wang$^{1*}$, Jiaming Lei$^{2*}$, Hai Lan$^{2}$, Arafat Al-Jawari$^{2}$, Xian Wei$^{3\dagger}$% <-this % stops a space
\thanks{*Equal technical contribution}% <-this % stops a space
\thanks{$^{1}$ Technical University of Munich,
        {\tt\small xihaowang2016@gmail.com}}%
\thanks{$^{2}$ Fujian Institute of Research on the Structure of Matter, Chinese Academy of Sciences, 
        {\tt\small scutwhite@gmail.com, lanhai09@fjirsm.ac.cn, arafatmmj@gmail.com}}%
\thanks{$^{3}$ East China Normal University,
        {\tt\small xian.wei@tum.de}}
\thanks{$\dagger$ Corresponding Author}
}

\begin{document}

\maketitle
\thispagestyle{empty}
\pagestyle{empty}
% \thispagestyle{plain}
% \pagestyle{plain}

%%%%%%%%%%%%%%%%%%%%%%%%%%%%%%%%%%%%%%%%%%%%%%%%%%%%%%%%%%%%%%%%%%%%%%%%%%%%%%%%
\begin{abstract}

% 室外点云识别在自动驾驶中的环境感知中扮演了及其重要的角色。在复杂的道路情况下，对目标朝向的精确识别为动态系统的预测与规划提供了不可或缺的信息，这大大提升了自动驾驶的安全性和可靠性（研究主题）。但是，现存的方法并没有专注于（研究目的与问题）
Outdoor 3D object detection has played an essential role in the environment perception of autonomous driving. In complicated traffic situations, precise object recognition provides indispensable information for prediction and planning in the dynamic system, improving self-driving safety and reliability. However, with the vehicle's veering, the constant rotation of the surrounding scenario makes a challenge for the perception systems. Yet most existing methods have not focused on alleviating the detection accuracy impairment brought by the vehicle's rotation, especially in outdoor 3D detection.  
%由此我们提出了DuEqNet,双等变网络。通过层级嵌套的结构，分别的提取了局部等级以及全局等级的等变特征。因此，我们的方法获得了丰富的方向信息以及显著提高了对目标朝向的准确性。(研究方法)
In this paper, we propose DuEqNet, which first introduces the concept of equivariance into 3D object detection network by leveraging a hierarchical embedded framework. The dual-equivariance of our model can extract the equivariant features at both local and global levels, respectively. For the local feature, we utilize the graph-based strategy to guarantee the equivariance of the feature in point cloud pillars. In terms of the global feature, the group equivariant convolution layers are adopted to aggregate the local feature to achieve the global equivariance. 
%在nuScenes数据集上的实验结果可以表明，该算法取得了50.49%的平均准确率(mAP)和0.3506的平均朝向误差(mAOE)，均比目前其他算法要更优。进一步实验也证明了该方法的泛化性（剩余实验以及优点和结论待补充）
In the experiment part, we evaluate our approach with different baselines in 3D object detection tasks and obtain State-Of-The-Art performance. According to the results, our model presents higher accuracy on orientation and better prediction efficiency. Moreover, our dual-equivariance strategy exhibits the satisfied plug-and-play ability on various popular object detection frameworks to improve their performance. 
% We test our approach with different baselines in 3D object detection tasks on the nuScenes dataset and obtain State-Of-The-Art performance, 50.49 mAP and 0.3506 mAOE. Further analysis demonstrates that dual-equivariance can capture enriched directional information and significantly improve object orientation accuracy. Moreover, the visualization provides an intuitive result to reveal that our method not only improves the accuracy of orientation but also has fewer invalid predictions. 
% 可视化结果显示所提算法比其他对比算法不仅具有更优的朝向预测，而且漏检误检情况更少。

\end{abstract}

%%%%%%%%%%%%%%%%%%%%%%%%%%%%%%%%%%%%%%%%%%%%%%%%%%%%%%%%%%%%%%%%%%%%%%%%%%%%%%%%
\section{INTRODUCTION}
\label{sec:1}
In recent years, autonomous driving techniques \cite{auto1,auto2,auto3} have achieved significant progress covering many scenarios, such as self-driving \cite{selfdriving}, robotaxis \cite{robotaxi}, and delivery robots \cite{robotdelivery}. As the core function of self-driving, precise 3D perception guarantees the safety and reliability of autonomous driving systems. The perception system generally receives multi-modality data in the complicated reality environment, including images from cameras, point clouds from LiDAR scanners, and high-definition maps \cite{survey1}. Under enriched input information, 3D object detection is one of the most important tasks to assist the automotive agent in understanding its surroundings comprehensively. Thus, several influence components are explored in terms of building an advanced perception performance in 3D object detection, such as object's shapes \cite{shape1,shape2}, sizes \cite{size1,size2} and locations \cite{location1, location2}.  
% As an important technique for autonomous driving systems, 3D object detection in outdoor environment has attracted widespread attention in industry as well as academics. 
% %
% However, in real and complicated driving scene, 3D object detection algorithm not only need to predict fast and precisely, but also predict the attributes of surrounded objects accurately, like orientation and shift.
% %展开说明一下为什么提升转角的预测精度有意义，可以考虑配图说明
% Precise orientation prediction can instruct an autonomous driving system to finish path planning, obstacle avoiding and other fundamental tasks.
% %
% Furthermore, in real scene, the variety of object orientations still poses a great challenge for detection methods to predict orientation exactly
\begin{figure}[t]
    \centering
    \includegraphics[scale=0.4]{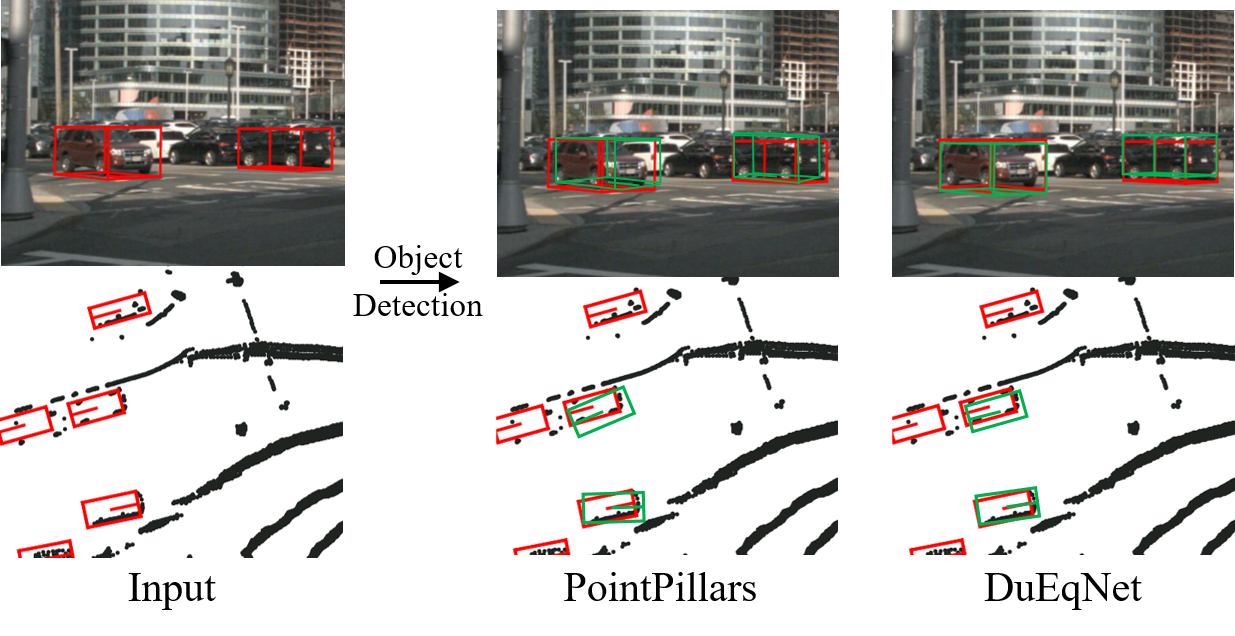}
    \caption{ The upper row is the image input, and the under row is the input from the LiDAR sensor. The left side of the figure shows the surrounding scenario, where the red box is the ground-truth bounding box. After the object detection process, the green box is the bounding box by prediction. The middle part describes the prediction result from the popular method of PoinPillars \cite{lang2019pointpillars}. Moreover, the right part presents the predicted result of our DuEqNet.}
    \label{fig:1}
\end{figure}
\unskip

With the development of 3D object detection techniques, more and more advanced methods are adopted for autonomous driving systems. However, in an actual driving situation, the vehicle needs to rotate its direction constantly, which causes a challenge for the perception systems toward the outdoor scenario. Influenced by the variation of the scenario orientation, the object detection accuracy suffered notable deterioration.
As illustrated in Figure \ref{fig:1}, without any method, the bounding boxes reveal poor quality during the vehicle rotation. 
Although object orientation is critical for 3D detection in outdoor scenarios, existing methods did not focus on improving the performance of the prediction of orientation. Rotation data augmentation is an indirect method to obtain a better orientation \cite{aug1}. In contrast, its expansive computation volume and unclear capture effect on orientation-related features indicate that rotation data augmentation is competent for gaining better object orientation prediction. 
% Thus, it's critical to improve the orientation prediction performance for 3D object detection method in intelligent driving scene.
% %
% In order to gain a better orientation prediction accuracy, currently, the most popular method is rotation data augmentation, i.e. randomly rotate the point cloud input to enrich the samples orientations, which can improve the generality of detection model to some extent and increase the orientation prediction accuracy. 
% %
% But, actually, rotation data augmentation enhances the model generality by enriching the samples, which does not promote the model to capture the orientation-related features and improve the ability of orientation prediction.

Therefore, inaccurate orientation prediction caused by consistent rotation challenges the ability of the existing 3D object detection methods. To address the mentioned problem, we propose a dual-equivariance 3D object detection network, called DuEqNet. We introduce a novel hierarchical embedded framework to extract the equivariant features at local and global levels. In particular, inspired by the embedding strategy in a directional message passing network\cite{directGNN}, we first propose a novel paradigm to extract local equivariance in pillars, which refers to the pillar-level rotation equivariance. Then, we introduce a lifting layer to generate the global equivariance in the pseudo feature map, which refers to the BEV (Bird's Eye View)-level rotation equivariance. The experiment results show that DuEqNet could be applied to the autonomous driving multimodel dataset \cite{nuscenes} and obtain the State-Of-The-Art (SOTA). 
%
% The emergence and popularity of PointPillars \cite{lang2019pointpillars} and subsequent PointPillars-based 3D object detection algorithms prove that 2D Convolution Neural Network (CNN) still has a strong vitality in 3D vision tasks.
% %
% However, CNNs have only translational invariance and can handle most vision tasks, such as classification and localization. 
% %
% Complex orientation prediction of 3D objects remains a huge challenge for CNNs. 
% %
% Therefore, in order to allow CNNs to identify and capture the orientation-related features of objects, and improve the performance, including orientation prediction and the overall average accuracy, we propose a 3D object detection algorithm based on equivariant convolution. Based on equivariance, the algorithm designs basic equivariant layers, such as space lifting layer and equivariant convolution layer.
% % and constructs an \textbf{\textit{Equivariant Feature Extraction Backbone (EFE-Backbone)}}. 
% With this backbone, the algorithm can pay more attention to orientation-related features and extract more helpful features.
%
Experiments on nuScenes dataset demonstrate that our proposed algorithm achieves a mean Average Orientation Error (mAOE) of $0.3506$ and a mean Average Precision (mAP) of $50.49 \%$, which is better than any of the currently used 3D object detection algorithms. 
% Moreover, through further experiments, our designed DuEqNet is applicable to most convolution-based 3D detection methods with considerable generalization.

The main contributions of this paper include:
\begin{itemize}
    \item To the best of our knowledge, we are the first to introduce the concept of dual-equivariance. It is a efficient approach that extracts the orientation-related feature in received perception information.  
    \item Based on the theory of equivariance, we elaborate a dual-equivariance framework for 3D object detection, named DuEqNet, which leverages a hierarchical embedded framework to extract the equivariant features at both local and global levels.
    % \item To evaluate our model, we conduct different object detection tasks on the nuScenes dataset to demonstrate the effectiveness of our method. We achieved the SOTA result compared with other methods. Besides, our dual-equivariance strategy exhibits the satisfied plug-and-play ability on various popular object detection frameworks. In the visualization analysis part, the target prediction result indicates that our method not only improves the accuracy of orientation but also has fewer invalid predictions.  
    \item In experiments, we achieve the SOTA result in different object detection tasks on the nuScenes dataset. Besides, our strategy exhibits the satisfied plug-and-play ability on various popular object detection frameworks. Moreover, the visualization of target prediction indicates that our method not only improves the orientation accuracy but also has fewer invalid predictions.
    
\end{itemize}

\begin{figure*}
    \centering
    \includegraphics[scale=0.52]{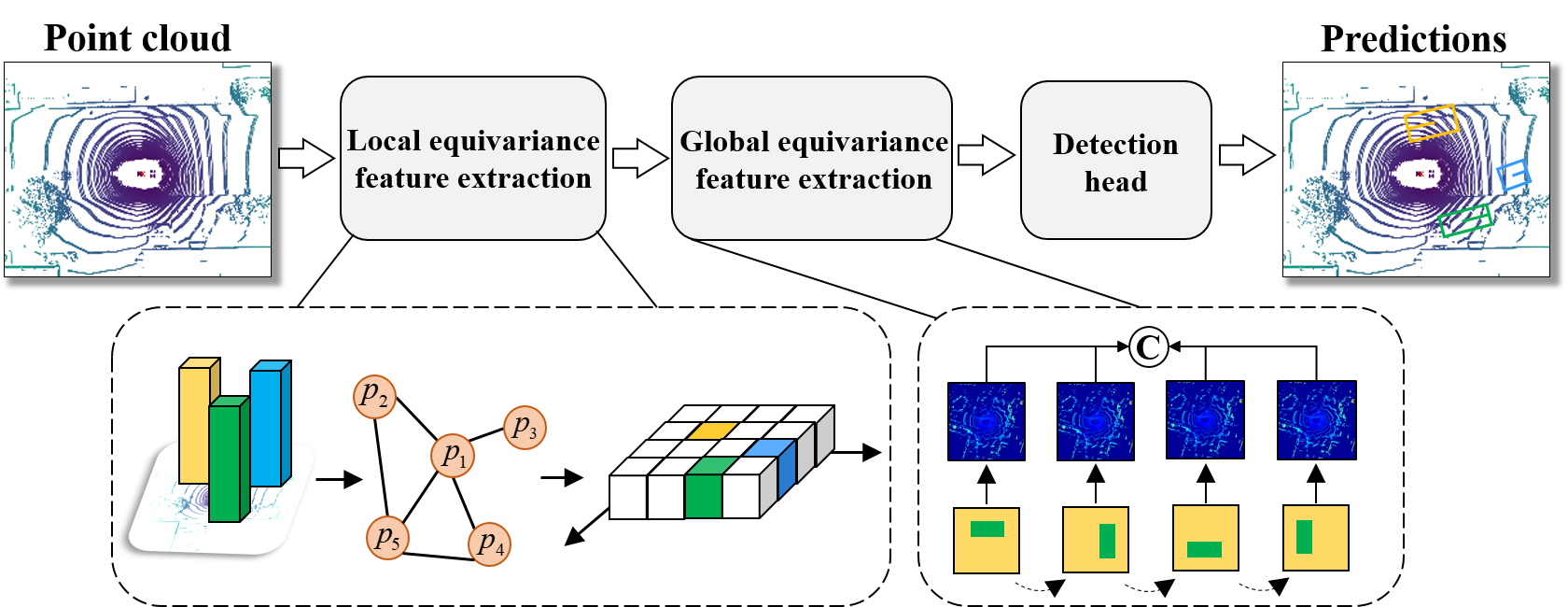}
    \caption{The overall framework of our proposed network. The DuEqNet can be divided into three parts: local equivariance feature extraction, global equivariance feature extraction and detection head.}
    \label{fig:2}
\end{figure*}
\unskip

\section{RELATED WORK}
\label{sec:2}

\subsection{3D object detection algorithm}
\label{sec:2.1}
% 主要介绍lidar的点云数据的3d目标检测算法，大多数3d目标检测算法都遵循着bottom-up设计，其中特征提取网络主要分为两部分，点云的特征提取部分。。。（这里介绍了voxel方法和graph）
In this section, we focus on the LiDAR-based 3D Object Detection algorithms, which directly take 3D point cloud data as input and predict 3D oriented bounding boxes (OBBs) to represent the objects in a scene \cite{mao20223d,lang2019pointpillars,pan20213d,yin2021centerpoint}. Most inherit the bottom-up design of deep learning models in which a backbone network is adopted to extract regional feature maps from the input data, with a subsequent detection head to propose candidate OBBs. Compared with images, point clouds are non-grid data that requires extra processing for feature extraction. VoxelNet \cite{zhou2018voxelnet} pioneered a voxelizing approach that establishes sparse voxel grids in which the points are encoded by a voxel feature encoding layer to extract features. This idea has inspired a series of follow-up researches \cite{DBLP:conf/aaai/DengSLZZL21_voxel_rcnn,DBLP:journals/pami/ShiWSWL21}. To improve the speed of voxelization, Lang et al. proposed pointpillars in which voxel size was limited to one among the vertical axis \cite{lang2019pointpillars}. Beyond voxelization, utilizing graph to encode point cloud is another promising methodology for feature extraction. Shi et al. designed Point-GNN to encode the point cloud into vertex in the graph for prediction of the objects \cite{DBLP:conf/cvpr/ShiR20_point_gnn}. SVGA-Net constructed the local complete graph and global KNN graph to serve as the attention mechanism for enhancing the extracted features \cite{DBLP:conf/aaai/HeWZZL22_svga_net}.

Another line of methods, the detection head is directly inspired by 2D detectors, and hierarchical architecture is adopted to enhance the performance of the detection head, while 3D objects are relatively small in the whole detection range. Second \cite{yan2018second} refined the SSD \cite{luo20203dssd} detection head and become the mainstream backbone in 3D object detection, later CenterPoint \cite{yin2021centerpoint} which is based on the CenterNet \cite{zhou2019objects_centernet} achieves state-of-the-art performance in the 3D object detection task on both nuScenes and Waymo \cite{sun2020waymo} datasets.

% However, these existing methods do not explicitly exploit object rotation equivariance in their models. Our method is grounded on this popular bottom-up modular design, and we focus on equipping these state-of-the-art detection models with object-level equivariance.

% % 这些工作虽然利用2d卷积取得了不错的应用表现，但是它们都没有关注待检测物体的朝向，这一项在自动驾驶中很重要的属性。因此，本文将在点云检测网络中引入等变性，促使网络更加关注朝向信息，从而提高其应用表现。
% Although the previous works perform well in real-time and precision through 2D convolution, they fail to focus on the orientation of detected objects, an essential property in autonomous driving. Therefore, we will introduce equivariance in point cloud detection network to induce them to pay more attention to orientation information and improve the overall performance.

\subsection{Equivariant Network}
\label{sec:2.2}

% 开创性工作，群等变卷积网络被Cohen 首次提出，具有旋转等变性。
As a pioneering work, the concept of an equivariant network was proposed in the group of equivariant convolution neural networks (G-CNN) by Cohen and Welling \cite{cohen2016group}.
% 不同于传统CNN，群卷积网络具有更高程度的权值共享，而且参数能够在群组之间共享。
Compared with normal ConvNet, G-CNN guarantees the rotation equivariance of extracted features under the group operation by a higher degree of weight sharing. Therefore, the expressivity of the equivariant network can be improved without significantly increasing the number of parameters. Depending on group equivariance, followed works extended the equivariance of network from discrete to continuous group. Li et al. \cite{li2018deep} applied a rotation transformation on the convolution kernel and proposed three novel convolutional layers. Finzi et al. \cite{finzi2020lieconv} proposed LieConv, which is theoretically equivariant to transformations from any Lie group, and constructed an equivariant CNN with higher generalization.

% 一直以来，GNN被认为是建模几何以及拓扑结构的有效方法，得益于深度学习的发展，它展现出了更加强大的生命力。尤其在处理几何和非结构化数据，如分子和点云。近年来，学者们深入探索发现这些数据隐式地蕴含了具有平移、旋转等变换对称性的几何图形。而尽管GNN是置换等变的，但是它本质上并不具有几何等变性。为了实现几何等变，大部分现有的工作都是通过修改GNN中的参数化函数来实现，包括消息传递和邻域聚合。这些设计的参数化函数通常只针对特定的变换，如旋转变换。例如 TFN在SE(3)群上实现平移和旋转的等变性。然而，这些方法都依赖于复杂的群论和几何知识，计算量大且难以扩展。

Moreover, the concept of equivariant network is also applied in graph neural networks (GNN), which have demonstrated their prominence in dealing with unstructured data, such as molecule and point clouds. These data implicitly incorporate geometric graphs which exhibit symmetries of translations, rotations, and other transformations. However, GNNs are always permutation equivariant but not inherently geometrically equivariant \cite{han2022geometrically}. To attain geometric equivariance, most works have been proposed to modify the parametric functions in GNN, including message passing and local aggregation. TFN \cite{thomas2018TFN} realizes translation and rotation equivariance on the group SE(3). To alleviate the burden of heavy computation consumption, DimeNet \cite{gasteiger2019directional} embeds the direction and distance between atoms as features to obtain rotationally equivariance with a simple model structure.

% 大量的工作都集中在分子、动力学系统以及室内点云分类中，少有工作将室外点云检测与等变性联系起来。对于pillar-based的点云检测算法，we consider每个pillar的内部都隐含一副图形，蕴含着特征信息。为了保持整个网络的简单以应用于自动驾驶，我们设计一个简单但有效的等变特征，并结合GNN实现等变pillar编码。
Although numerous applications of equivariant networks have been confined to molecules, physical dynamics simulation, and indoor point cloud classification, little work established the link between outdoor point cloud object detection and equivariant network due to the massive scale of data. For 3D object detection tasks, the points in each pillar imply a graph containing geometric information (like orientation-related information), but most approaches do not explicitly exploit object rotation equivariance in their models. To tackle this problem, we propose DuEqNet, which focuses on attaching object-level equivariance to the object detection model by designing a simple but effectively equivariant feature through equivariant pillar encoding with GNN. 
% 不同于理论研究，我们专注于群等变卷积网络对于方向特征的提取。出于容易实现的考虑，我们将简单的群等变卷积嵌入到3d目标检测算法中，作为特征提取网络，以增强算法对于目标朝向的识别能力，从而提高整体算法性能。
% Unlike theoretical research, we focus on the extraction of directional features by equivariant convolution. For ease of implementation, we simply embed an equivariant convolution network in the 3D object detection algorithm as a feature extraction backbone to enhance the capability of object orientation identification, resulting in improved performance. 

%%%%%%%%%%%%%%%%%%%%%%%%%%%%%%%%%%%%%%%%%%
\section{Proposed Method}
\label{sec:3}
In this section, a brief overview of our network and the preliminary about equivariance representation are given first. Then, we describe the strategy of extracting the local and global equivariance features individually. Finally, we introduce the connection of two levels of features. 

\subsection{Overview and Preliminary}
\label{sec:3.1}
% 为了应对真实驾驶场景中车辆不断转向对识别系统带来的挑战，我们提出了双等变网络
To cope with the challenge induced by scenario rotation,  we propose the hierarchical embedded framework to extract equivariance features. The architecture of the network is presented in Figure \ref{fig:2}. The architecture of the network can be represented as:
\begin{equation}
\label{}
    \mathcal{B}={f}_{det}(\mathbf{{G}_{e}}(\mathbf{{L}_{e}}(x^{p_i}_{\cdot}))),
\end{equation}
where $x^{p_i}_{\cdot}\in\mathbb{R}^{\alpha}$ denotes each point with in pillar $p_i\in\mathbb{R}^{\beta}$. In detail, $\alpha$ is the initial feature of the input point, and $\beta$ is the embedded pillar's vector dimension. $\mathbf{L_e}$ and $\mathbf{G_e}$ represent the local and global equivariance function, respectively. $f_{det}$ is a detection head to detect and regress 3D bounding boxes $\mathcal{B}$. In terms of the equivariance feature, following the definition of group equivariance \cite{cohen2016group}, a function $f: X \rightarrow Y$, where $X$ and $Y$ are two homomorphism spaces, is defined as being equivariant if $ f(\varphi^{X}_{g}(x))=\varphi^{Y}_{g}(f(x))$ with the group action $g$ in the input space $\varphi^{X}_{g}$ and output space $\varphi^{Y}_{g}$ \cite{gasteiger2019directional}. The pillars-level and BEV-level features operation could be considered the group action between two homomorphism spaces. 

% 双等变结构主要体现在函数L_e和G_e. 在等变pillar-encoder中，L_e在pillar内部实现局部等变。 在等变特征提取骨干网络中，G_e在pillar之间实现全局等变，以提取丰富特征。
% The dual-equivariance structure is mainly reflected in function $\mathbf{L_e}$ and $\mathbf{G_e}$. In equivariant pillar-encoder, $\mathbf{L_e}$ implements local equivariance inside pillars with equivariant GNN. In equivariant feature extraction backbone (\textbf{\textit{EFE-Backbone}}), $\mathbf{G_e}$ realizes global equivariance between pillars for rich features extraction.

%%%%%%%%%%%%%%%%%%%%%%%%%%%%%%%%%%%%%%%%%%%%%%%
\subsection{Local Equivariance Feature Extraction}
\label{sec:3.2}
To capture the local geometric information, we apply the pillars as our data representation for 3D object detection. Compared with the voxels, pillars have an unlimited size in the vertical direction. Inspired by the message passing strategy \cite{messagepassing} in the graph network, we consider each pillar as the subgraph from input data. Besides, each 3D input point in the pillar is similar to the node $u\in \mathcal{V}$ in the complete graph $\mathcal{G}\in (\mathcal{V},\mathcal{E})$, where $\mathcal{V}$ is the vertex and $\mathcal{E}$ is the edge. As the original 3D input does not satisfy the rotation equivariance, we require the embedding for each node $u_{i}$ and neighbor $u_{j}$ by the same learned filter $W(u_{j}-u_{i})$ with the distance of each neighboring input. Hence, the update of passing message $m_{ji}$ between sampled input and its neighborhood $\mathcal{N}_{i}$ with $\sigma$ nodes is defined as: 
\begin{equation}
\label{equ:2}
    m_{j i}^{(l+1)}=f_{\text {update }}\left(m_{j i}^{(l)}, \sum_{\zeta \in \mathcal{N}_{i} } f_{\text {agg}}\left(m_{\zeta j}^{(l)}, d_{\mathrm{RBF}}^{(j i)}\right)\right),
\end{equation}
where $d_{\mathrm{RBF}}^{(j i)}\in\mathbb{R}^{\alpha}$ denotes the pair-wise point distance represented from the radial basis function, $m_{\zeta j}=W(u_{\zeta}-u_{j})=W(||u_{\zeta}-u_{j}||_{2})\in\mathbb{R}^{\sigma -1}$ and $f_{agg}$ denotes the linear aggregation function. According to the equivariance of the function, we message update function is presented as: 
\begin{equation}
    f_{\text{update}}(\varphi^{(l)}_{r}(x^{p_i}_{\cdot}))=\varphi^{(l+1)}_{r}(f_{\text{update}}(x^{p_i}_{\cdot})),
\end{equation}
 where $r$ denotes the rotation operation. Thus, we consider that the message passing paradigm $\mathbf{L_e}$ in the pillar satisfies the equivariant constraint, which means that the local equivariance feature is extracted in the corresponding pillar.

\subsection{Global Equivariance Feature Extraction}
\label{sec:3.3}
% 为了从equivariant pillar-encoder中的pillar局部等变信息中提取全局等变特征，我们构造了等变特征提取骨干网络 (i.e. G_e)，它能够实现全局等变。依据群等变卷积的概念，G_e可以表示如下：
After extracting the local equivariance features, we construct the module $\mathbf{G_e}$ to realize global equivariance between pillars. According to the concept of group equivariant convolution, $\mathbf{G_e}$ can be shown as:
\begin{equation}
\label{equ:3}
    \mathbf{G_e}(x)=\text{ReLU}(\text{BN}_e(\Psi_e(x))),
\end{equation}
% Psi_e包括空间提升函数L和群卷积C，BN_e是满足群定义的bn。
where $\Psi_e$ includes space lifting function $\mathcal{L}$, group convolution $\mathcal{C}$ (including convolution and transposed convolution), x is the pseudo feature map aggregated via local feature extraction, and $BN_e$ is the batch normalization which satisfies the definition of group. We explain the reason for applying specific batch normalization in the Section \ref{sec:3.4}.

%%%%%%%%%%%%%%%%%%%%%%%%%%%%%%%%%%%
% 空间提升函数L能够将输入特征图从空间X映射到一个更大的空间Y。在空间Y中，所有的操作将严格满足P4群的约束。将卷积核定义为 \Psi \in X，输入特征x\in X，则函数L则以由方程3定义：
With the purpose of achieving the global rotation equivariance, we build the space lifting function $\mathcal{L}$. It maps the space $X$ of pseudo feature to a larger homomorphism space $Y$. 
% All operations in space Y will satisfy the constraint defined according to the group $P_4$.Represent the convolution kernel $\Psi\in X$, input feature map $x\in X$, then 
The lifting convolution $\Psi\star x$ is defined as follows:
\begin{equation}
\label{equation:L}
\begin{aligned}
    \relax[\Psi\star x](t,r) & :=\sum_{\text{p}\in\mathbb{Z}^2}\Psi((t,r)^{-1}\text{p})x(\text{p}) \\
    & =\sum_{\text{p}\in\mathbb{Z}^2}\Psi(r^{-1}(\text{p}-t))x(\text{p}),
\end{aligned}
\end{equation}
where $x(\text{p})$ denotes the value of a pixel point $\text{p}$ in pseudo feature map $x$, $\Psi$ is the group convolution kernel in space $X$, and $(t,r)$ denotes the element in group $P_4$. The group $P_4$ is the symmetry group that collects all combinations of translations $t$ and 90-degree rotations $r$ about any square grid's center \cite{cohen2016group}. Unlike the regular convolution operator, the group convolution consists of the coupled space $\mathbb{R}^{2}\times S^{1}$ of translations in 2D plane space $\mathbb{R}^{2}$ and rotations in 1D spherical space $S^{1}$. Thus, the representation of the group convolution could be expressed as 
\begin{equation}
    (k * f)(\hat{x})=(\mathcal{L}_{r}^{S^{1}\rightarrow \mathbf{L}_{2}(\mathbb{R}^{2})}\mathcal{L}_{r}^{\mathbb{R}^{2}\rightarrow \mathbf{L}_{2}(\mathbb{R}^{2})}k,f)_{\mathbb{L}_{2}(\mathbb{R}^{2})},
\end{equation}
where $\mathbb{L}$ is the left representation, $k$ is the kernel of the convolution, and $\hat{x}=(t,r)$ with the group elements in group $P_{4}$. 
% i.e., to perform a rotation in the cyclic group $r\in C_4$ first and a plane translation of $t$ on a point $\text{p}\in\mathbb{Z}^2$. 
Due to the road traffic scene, most of the vehicle's rotation is 90-degree. Consequently, we decide to choose the rotation operator $r$ belongs to the cyclic group $C_{4}$. In terms of the definition of equivariance, the output of function $\mathcal{L}$, $\mathcal{C}$ naturally satisfy:
\begin{equation}
\label{}
\begin{aligned}
    \relax[(t,r)\, y](\text{p},s) & =y((t,r)^{-1}\, (\text{p},s)) \\
    & =y(r^{-1}(\text{p}-t),r^{-1}s),
\end{aligned}
\end{equation}

where $s$ denotes the additional rotation element of the input feature. For a concise representation, let ${g}=(t,r)\in P_4$, then the rotation equivariance of this layer is presented as follows:
\begin{equation}
\begin{aligned}
\label{}
\relax
    [\Psi\star(\mathcal{R}x)](g) & = \sum_{y\in\mathbb{Z}^2}\sum_{k}x_k(\mathcal{R}^{-1}y)\Psi_k(g^{-1}y) \\
    & \overset{y=\mathcal{R}y}{=} \sum_{y\in\mathbb{Z}^2}\sum{k}x_k(\Psi_k(g^{-1}\mathcal{R}y) \\
    & = \sum_{y\in\mathbb{Z}^2}\sum{k}x_k(y)\Psi_k((\mathcal{R}^{-1}g)^{-1}y) \\
    & = [\mathcal{R}(\Psi\star x)](g),
\end{aligned}
\end{equation}
where $\mathcal{R}$ denotes rotation, and $x$ and $\Psi$ have the same definition as above. Based on the concept of group equivariance, we construct equivariant convolution and transposed convolution to extract rich features following function $\mathcal{L}$.

\subsection{Joint of Dual-Equivariance Framework}
\label{sec:3.4}
\begin{figure}
    \centering
    \includegraphics[scale=0.38]{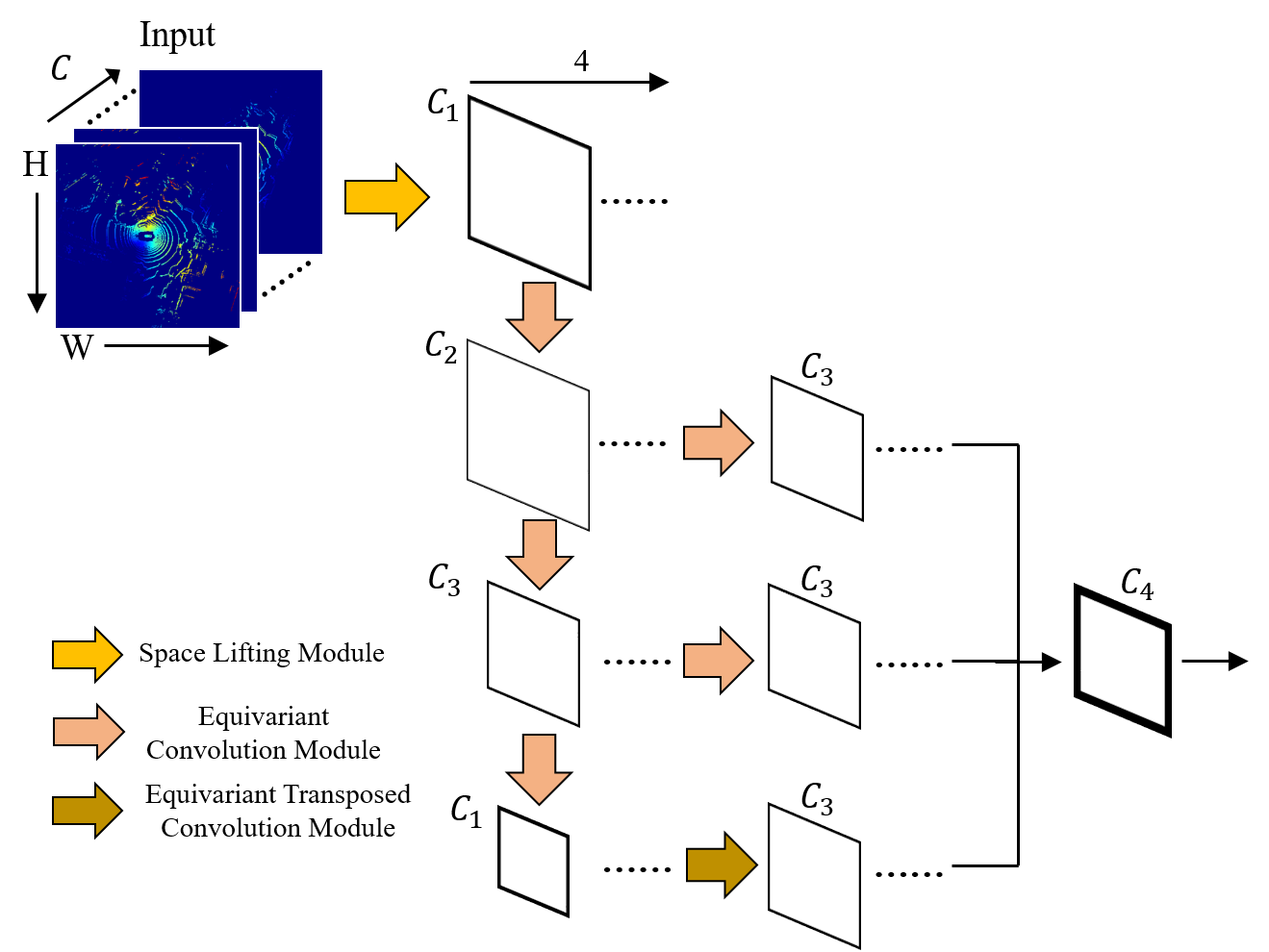}
    \caption{The framework of feature extraction module used in our algorithm. After space lifting, convolution and transposed convolution are used for feature extraction. Finally, stacking three feature maps for final output.}
    \label{fig:6}
\end{figure}

\textbf{Equivariant batch normalization} $BN_e$: Batch Normalization (BN) can stabilize the intermediate layers, accelerate the convergence of the network, and suppress overfitting to some extent. In order to embed BN into backbone, we base on the definition of group and implement $BN_e$. Considering the process of BN, it only change the data distribution of the output data, but not the space where the output is located. 
% So, the process of $\mathcal{B}$ should be illustrated in Algorithm \ref{algorithm:1}.
% \begin{algorithm}[htb]
% \caption{The implementation of equivariant batch normalization $BN_e$}
% \label{algorithm:1}
% \setstretch{1.5}
% \begin{algorithmic}[]
% \STATE {\bf Input:} $y\in\mathbb{R}^{B\times C\times 4\times H\times W}$, denotes the output of $\mathcal{C}$ or $\mathcal{T}$ \\
% \STATE {\bf Output:} $y'\in\mathbb{R}^{B\times C\times 4\times H\times W}$, denotes the output of $\mathcal{B}$ \\
% \STATE For each $y'_i=y'[\cdot,\cdot,i]\in\mathbb{R}^{B\times C\times H\times W},i=\{0,1,2,3\}$: \\
%     \STATE \ \quad 1. Calculate the mean of $y'_i$: $\mu_{y'_i}=\frac{1}{b}\sum_{j=1}^B y'_{ij}$ \\
%     \STATE \ \quad 2. Calculate the variance of $y'_i$: $\sigma_{y'_i}^2=\frac{1}{B}\sum_{j=1}^B (y'_{ij}-\mu_{y'_i})^2$\\
%     \STATE \ \quad 3. Normalize $y'_i$: $\widehat{y'_i}=\frac{y'_i-\mu_{y'_i}}{\sqrt{\sigma^2_{y'_i}+\epsilon}}$\\
%     \STATE \ \quad 4. Learnable parameters (offset $\beta$ and scaling $\gamma$): $y'_i=\gamma\widehat{y'_i}+\beta$ \\
% \STATE Stacking all the $y'_i$ to get the final output $y'$ \\
% \end{algorithmic}
% \end{algorithm}
%%%%%%%%%%%%%%%%%%%%%%%%%%%%%%%%%%%%%%%%%%%%%%
% EFE-Backbone：综上所述，我们可以通过组合和堆叠上述函数，从而搭建出一个完成满足旋转等变性的特征提取网络，包括空间提升L，等变卷积C（卷积或反卷积），激活函数以及BN_e，如图所示。
% \textbf{Equivariant feature extraction backbone}: 
By combining and stacking the above functions, we can build a rotationally equivariant feature extraction module (Figure \ref{fig:6}). It consists of lifting $\mathcal{L}$, convolution $\mathcal{C}$ (convolution and transposed convolution), activation and batch normalization $BN_e$.

%%%%%%%%%%%%%%%%%%%%%%%%%%%%%%%%%%%%%%%%%%%%%%%%%

\textbf{Detection head}: Considering the diversity of object orientations in 3D scenes and that many orientations are not aligned parallel to the coordinate axes, we adopt a center-based detection head \cite{yin2021centerpoint} to represent object better and predict object orientation more precisely. In our model, the object is described as points and regress orientation, size, velocity and other attributes.

\begin{table*}[h!]
\caption{3D detection mAP($\%$) and NDS on nuScenes validataion set.}
\label{tab:1}
\begin{center}
\begin{tabular}{c|cccccccccc|c|c}
\hline
\bfseries Method & \bfseries car & \bfseries peds. & \bfseries barr. & \bfseries traf. & \bfseries truck & \bfseries bus & \bfseries trail. & \bfseries cons. & \bfseries motor. & \bfseries bicy. & \bfseries mAP & \bfseries NDS \\
\hline
SARPNET \cite{ye2020sarpnet}             & 59.9 & 69.4 & 38.3 & 44.6 & 18.7 & 19.4 & 18.0 & 11.6 & 29.8 & 14.2 & 32.4 & 48.4 \\
PointPillars \cite{lang2019pointpillars} & 78.7 & 61.2 & 41.4 & 18.9 & 37.2 & 49.7 & 26.2 & 6.56 & 20.2 & 0.85 & 34.1 & 49.9          \\
WYSIWYG \cite{Hu_2020_WYSIWYG}           & 80.0 & 66.9 & 34.5 & 27.8 & 35.8 & 54.1 & 28.5 & 7.50 & 18.5 & 0.0 & 35.4 & - \\
InfoFocus \cite{wang2020infofocus}       & 77.9 & 63.4 & 47.8 & 46.5 & 31.4 & 44.8 & 37.3 & 10.7 & 29.0 & 6.1 & 39.5 & -          \\
SSN \cite{zhu2020ssn}                    & 81.0 & 66.2 & 49.6 & 18.7 & 45.0 & 53.0 & 26.1 & 10.6 & 41.0 & 20.5 & 41.2 & 54.8          \\
3DSSD \cite{luo20203dssd}                & 81.2 & 70.2 & 47.9 & 31.1 & 47.2 & 61.4 & 30.5 & 12.6 & 36.0 & 8.63 & 42.7 & 56.4          \\
Free-anchor3d \cite{zhang2019freeanchor} & 81.2 & 74.4 & 52.7 & 41.4 & 39.3 & 48.0 & 30.9 & 10.2 & 43.5 & 18.0 & 44.0 & 55.0          \\
PointPainting \cite{vora2020pointpainting}&77.9 & 73.3 & 60.2 & \textbf{62.4} & 35.8 & 36.1 & \textbf{37.3} & \textbf{15.8} & 41.5 & \textbf{24.1} & 46.4 & 58.1          \\
CenterPoint \cite{yin2021centerpoint}    & 83.9 & 77.3 & 60.1 & 50.5 & 50.2 & 62.0 & 32.7 & 10.5 & \textbf{45.4} & 16.4 & 48.9 & 59.6          \\
\hline
DuEqNet                                   & \textbf{84.2} & \textbf{78.9} & \textbf{61.3} & 56.6 & \textbf{52.4} & \textbf{64.6} & 32.4 & 13.0 & 45.3 & 16.3 & \textbf{50.5} & \textbf{60.6} \\
\hline
\end{tabular}
\end{center}
\end{table*}
\unskip

\begin{table*}[h!]
\caption{Per class AOE on nuScenes validataion set.}
\label{tab:2}
\begin{center}
\begin{tabular}{c|ccccccccc|c}
\hline
\bfseries Method & \bfseries car & \bfseries peds. & \bfseries barr. & \bfseries truck & \bfseries bus & \bfseries trail. & \bfseries cons. & \bfseries motor. & \bfseries bicy. & \bfseries mAOE($\downarrow$) \\
\hline
Free-anchor3d \cite{zhang2019freeanchor} & 0.1618 & \textbf{0.3632} & \textbf{0.0550} & 0.2688 & 0.2808 & 0.5437 & 1.4663 & 0.6709 & 0.9576 & 0.5298 \\
PointPillars \cite{lang2019pointpillars} & 0.1569 & 0.4265 & 0.0560 & 0.2117 & 0.3295 & 0.3751 & 1.5133 & 0.7792 & 0.8593 & 0.5231 \\
SSN \cite{zhu2020ssn}                    & 0.1531 & 0.4103 & 0.0523 & 0.1653 & 0.2022 & \textbf{0.3745} & 1.2489 & 0.5336 & 0.7797 & 0.4355 \\
CenterPoint \cite{yin2021centerpoint}    & 0.1642 & 0.4108 & 0.0799 & 0.1532 & 0.0791 & 0.5340 & 0.9745 & 0.4279 & 0.6416 & 0.3850 \\
\hline
DuEqNet                                  & \textbf{0.1494} & 0.4013 & 0.0812 & \textbf{0.1423} & \textbf{0.0475} & 0.5178 & \textbf{0.8553} & \textbf{0.3622} & \textbf{0.5980} & \textbf{0.3506} \\
\hline
\end{tabular}
\end{center}
\end{table*}
\unskip

\section{Experiments and Analysis of Results}
\label{sec:4}

\subsection{Dataset and Evaluation Metrics}
\label{sec:4.1}
We conducted all the experiments in this paper on the nuScenes dataset. As one of the most popular large public datasets in autonomous driving, the nuScenes dataset is equipped with a 32-line LiDAR deployed on the top and middle of the vehicle. It contains a total of $40000$ key frames collected from different scene locations, including $28130$ training samples and $6019$ validation samples with 23 categories of labelled objects, like cars, pedestrians, and cyclists. While for 3D object detection task, we need to detect objects of 10 classes.

For 3D object detection, the most commonly used evaluation metric is the average precision (AP), which evaluates the center distance between the prediction and the ground truth in the bird-eye view in the nuScenes dataset. 
% For each category, its final AP is calculated based on the APs under four distance thresholds, shown in Equation \eqref{equation:5}:
% \begin{equation}
% \label{equation:5}
%     \text{AP}_c=\frac{1}{|\mathbb{D}|}\sum_{d\in \mathbb{D}}\text{AP}_d,
% \end{equation}
% where $\mathbb{D}=\{0.5,1.0,2.0,4.0\}$ represents distance thresholds of $0.5$m, $1.0$m, $2.0$m and $4.0$m, respectively. And $\text{mAP}$ is computed by: $\text{mAP}=\frac{1}{10}\sum_{c\in\mathbb{C}}\text{AP}_c$.

Besides AP, nuScenes also measures a series of True Positive metrics (TP metrics) to assess the center distance, size, orientation, velocity and classification deviation between prediction and ground truth. Noted that the mean Average Orientation Error (mAOE) is the one of the importance metric in our experiments, to validate the effectiveness of our Dual-Equivariance structure. Moreover, in order to consider the mAP and all TP metrics, nuScenes proposed the nuScenes detection score (NDS) (Equation \eqref{equation:6})

\begin{equation}
\label{equation:6}
    \text{NDS}=\frac{1}{10}[5\text{mAP}+\sum_{\text{mTP}\in\mathbb{TP}}(1-\min{(1,\text{mTP})})].
\end{equation}

\subsection{Setting in Training and Inference}
\label{sec:4.2}

As described in Section \ref{sec:4.1}, this paper trains a 10-class detection network and evaluate its AP, TP and NDS. We focus on mAOE to evaluate and analyze the effectiveness of our proposed algorithm for improving the accuracy of orientation prediction.
In all our experiments, we set the detection range of LiDAR point cloud to [-51.2m, 51.2m] for \textit{X} and \textit{Y} axis, and [-5m, 3m] for \textit{Z} axis.
Moreover, following the official pre-processing baseline for point cloud data, we aggregate the key frame with $9$ consecutive frames before feeding into the network for the more dense sample frame.

We perform data augmentations on each training, including random rotation, random scaling and random flip along the \textit{X} or \textit{Y} axis. In validation and testing, no data augmentation is performed. For a fair comparison, all experiments were conducted on the whole dataset with training 24 epochs on the same machine of GeForce RTX 3090 GPU. The networks were trained and validated on eight 3090 GPUs with the AdamW algorithm with 0.01 weight decay for optimization.

\subsection{Results and analysis}
\label{sec:4.3}
In this section, we exhibit the results of our DuEqNet and compare them to other popular methods. Following tables obey these abbreviations: pedestrian (peds.), barrier (barr.), traffic cone (traf.), trailer (trail.), construction vehicle (cons.), motorcycle (motor.) and bicycle (bicy.).

\textbf{Precision:} Firstly, we compare the precision of our network with other popular detection methods, including lidar-based and fusion-based methods. As shown in Table \ref{tab:1}, DuEqNet outperforms other methods in terms of mAP and NDS. Compared with the second best method CenterPoint, our method shows an absolute improvement of $1.6\%$ and $1.0\%$ in mAP and NDS. It means that DuEqNet not only locates potential objects precisely but also predicts the attribute of objects more comprehensively.
%是否要再多一些分析和原因？
Furthermore, DuEqNet gains the highest AP results on most classes, such as cars and pedestrians. For traffic cones, trailers, construction vehicles, and bicycles, PointPainting gets the best APs, which are slightly higher than our DuEqNet. One reasonable reason is that PointPainting accepts LiDAR point cloud data and camera images as input, while DuEqNet only employs point cloud data. Images can offer rich color and texture information, improving the overall performance of the fusion model. DuEqNet obtains $45.3\%$ AP for motorcycles, which is just $0.1\%$ lower than CenterPoint. It has a comparable capability to CenterPoint for detecting motorcycles.

\textbf{Orientation:} In addition, to verify our method for orientation prediction, we evaluate the AOE of each class and the mAOE in Table \ref{tab:2}. Our method DuEqNet achieves the lowest mAOE and performs best in most classes. More specifically, it surpasses CenterPoint, the most popular method, with a relative improvement of $8.9\%$. The significant improvement of AOE is supported by our visualization results in Figure \ref{fig:10}(a), where our methods show a better orientation prediction.

\textbf{Ablation study:} We provide ablation studies to assess the effectiveness of the proposed dual-equivariance structure. Recall that our dual-equivariance structure consists of two parts: local and global equivariance feature extraction ($\mathbf{L_e}$ and $\mathbf{G_e}$ for concise representation in tables, respectively). For baseline (remarked as \textbf{Idx. 1} in Table \ref{tab:3}), we adopt the same method as PointPillars \cite{lang2019pointpillars} in pillar encoding and replace global equivariance feature extraction with popular convolution backbone \cite{yin2021centerpoint}, \cite{lang2019pointpillars}. 

In Table \ref{tab:3}, with dual-equivariance structure, our DuEqNet (\textbf{Idx. 4}) achieves the best results including NDS, mAP and mAOE. Concretely, compared to the baseline, we attain absolute progress by $1.07$, $1.59$ concerning NDS and mAP and $8.94\%$ orientation error reduction. Note that, from the results of baseline and method \textbf{3}, the global equivariance feature extraction substantially lowers the mAOE from $0.3850$ to $0.3598$. It shows a more powerful effect on orientation than the local equivariance feature extraction. This can be attributed to the size of objects: most objects occupy several pillars in point cloud. The global equivariance feature extraction can capture the relationship between pillars which is beneficial for orientation regression.

\begin{table}
\caption{Ablation studies on nuScenes validation set}
\label{tab:3}
\centering
\begin{tabular}{c|c|c|c|c|c}
\hline
\bfseries Idx.	& $\mathbf{L_e} $	& $\mathbf{G_e}$ & \bfseries NDS & \bfseries mAP & \bfseries mAOE \\
\hline\hline
1 & \ding{56} & \ding{56} & 59.55           & 48.90            & 0.3850\\
2 & \ding{52} & \ding{56} & 59.83           & 49.28            & 0.3821\\
3 & \ding{56} & \ding{52} & 60.31           & 50.17            & 0.3598\\
4 & \ding{52} & \ding{52} & \textbf{60.62}  & \textbf{50.49}   & \textbf{0.3506}\\
\hline
\end{tabular}
\end{table}
\unskip

% From the results of the ablation experiments, both the rotation data augmentation (RDA) and our proposed \textbf{\textit{EFE-Backbone}} can promote the mAP, mAOE and NDS. Comparing the performance of algorithm \textbf{2} and \textbf{3}, without RDA, a comparable mAP performance can be obtained with our proposed \textbf{\textit{EFE-Backbone}}. While a larger improvement in orientation prediction (mAOE decreases from $0.3850$ to $0.3694$) proves the effectiveness of \textbf{\textit{EFE-Backbone}}, which can enhance the algorithm for orientation identification and prediction.

% Moreover, the algorithm which combines RDA and \textbf{\textit{EFE-Backbone}} (algorithm \textbf{4}) achieves the best result. One reason is that the data augmentation can increase the generalization of the algorithm. More importantly, our proposed \textbf{\textit{EFE-Backbone}} enhances the ability of feature representation by mapping the features to a more complex space, thus improving the performance of the algorithm.

%
\textbf{Generalization:} In order to investigate the generalization of our dual-equivariance structure, we conduct a further experiment. We simply replace the parts of pillar encoding and feature extraction with our dual-equivariance structure in the popular 3D object detection methods.

As is shown in Table \ref{tab:4}, we compare four methods, including PointPillars, SSN, Free-anchor3d and CenterPoint, on nuScenes validation set. With our dual-equivariance structure, all methods receive a considerable increment in mAP and NDS, demonstrating the generalization of our dual-equivariance structute.

\begin{table}
\caption{the generalization of dual-equivariance structure on nuScenes validation set}
\label{tab:4}
\centering
\begin{tabular}{c|c|c|c}
\hline
\multicolumn{2}{c|}{\bfseries Method} & \bfseries mAP & \bfseries NDS \\
\hline
\multirow{2}*{PointPillars \cite{lang2019pointpillars}} & w.o. & 34.09 & 48.95\\
~ & w. & 34.59(+0.50) & 49.40(+0.45) \\
\hline
\multirow{2}*{SSN \cite{zhu2020ssn}} & w.o. & 41.17 & 54.56\\
~ & w. & 42.57(+1.4) & 54.77(+0.21) \\
\hline
\multirow{2}*{Free-anchor3d \cite{zhang2019freeanchor}} & w.o. & 43.96 & 54.98 \\
~ & w. & 45.55(+1.62) & 56.07(+1.09) \\
\hline
\multirow{2}*{CenterPoint \cite{yin2021centerpoint}} & w.o. & 48.90 & 59.55 \\
~ & w. & 50.36(+1.46) & 60.50(+0.95) \\
\hline
\end{tabular}
\end{table}
\unskip

% \begin{figure}
%     \centering
%     \includegraphics[scale=0.2]{fig/fig9.png}
%     \caption{mAP and mAOE comparison of different algorithms with or without \textbf{\textit{EFE-Backbone}}.}
%     \label{fig:9}
% \end{figure}

\subsection{Visualization Analysis}
\label{sec:4.5}
\begin{figure}[H]
    \centering
    \includegraphics[scale=0.60]{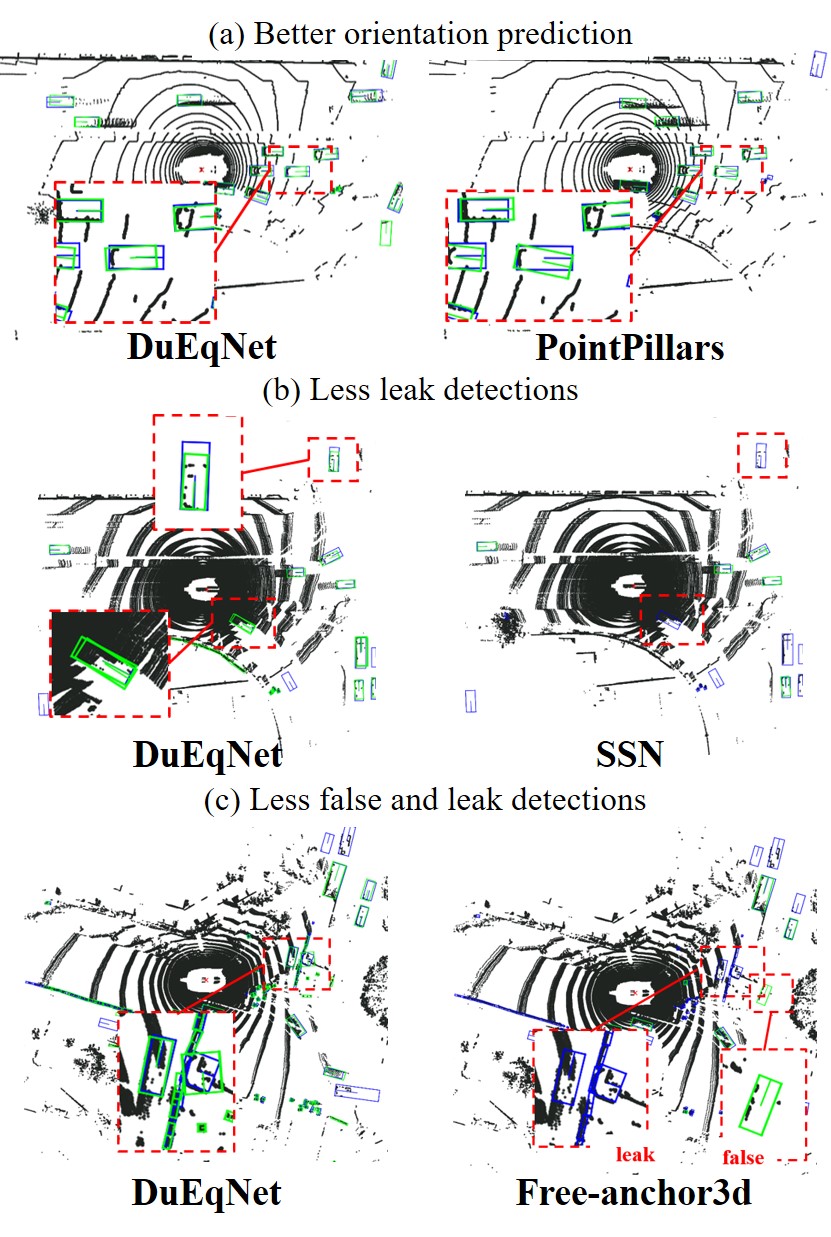}
    \caption{Visualization of algorithms' prediction. Our detection is shown in the left column, while others is in the right. We compare our DuEqNet with other methods in three aspects: orientation prediction, leak prediction, and false prediction.}
    \label{fig:10}
\end{figure}

Setting the visual range of the x-axis and y-axis as $[-40m,40m]$, we implement the visualization analysis of the detection results of \textbf{DuEqNet} and other detection methods under bird's eye view. Presenting in Figure \ref{fig:10}. The blue bounding boxes represent the ground truth, and the green ones mean prediction boxes. The line in the boxes indicates the direction of the objects. From Figure \ref{fig:10}, \textbf{DuEqNet} can obtain better orientation prediction and effectively reduce the occurrence of leak and false detection. Visualization results demonstrate the validness of dual-equivariance feature extraction, which is beneficial to orientation prediction and performance improvement.

\unskip

\section{Conclusion}
\label{sec:5}
% 
% In natural driving scenes, because of the object orientations' diversity, enhancing detection algorithms' ability to recognize the orientation can significantly improve their performance. We proposed a 3D object detection algorithm based on equivariance, including equivariant pillar-encoder and equivariant feature extraction backbone. The experiments demonstrate the effectiveness and power of these two modules, which can help to pay more attention to orientation-related features. Moreover, the further experiments prove that our proposed modules have good generalization.
In this paper, we present a dual equivariance network for outdoor 3D object detection named DuEqNet, which employs a hierarchical embedded framework to extract the equivariance features at local and global levels. Through this efficient approach, the challenge brought by scenario rotation in autonomous driving is effectively mitigated. We demonstrate that our dual-equivariance concept advances the accuracy of object detection. Moreover, our network has generalization for other methods to improve their performance. In the days of autonomous driving development, the concept of dual equivariance introduced by our network provides a fresh perspective on enhancing self-driving safety.   
% 
% We link equivariance with 3D object detection


%%%%%%%%%%%%%%%%%%%%%%%%%%%%%%%%%%%%%%%%%%%%%%%%%%%%%%%%%%%%%%%%%%%%%%%%%%%%%%%%

%%%%%%%%%%%%%%%%%%%%%%%%%%%%%%%%%%%%%%%%%%%%%%%%%%%%%%%%%%%%%%%%%%%%%%%%%%%%%%%%



\section*{ACKNOWLEDGMENT}
This work was partially supported by 'Fujian Science \& Technology Innovation Laboratory for Optoelectronic Information of China' (Grant 2021ZZ120), 'FuJian Science and Technology Plan' (Grant 2021T3003) and 'QuanZhou Science and Technology Plan' (Grant 2021C065L).
% The preferred spelling of the word ÒacknowledgmentÓ in America is without an ÒeÓ after the ÒgÓ. Avoid the stilted expression, ÒOne of us (R. B. G.) thanks . . .Ó  Instead, try ÒR. B. G. thanksÓ. Put sponsor acknowledgments in the unnumbered footnote on the first page.

% %%%%%%%%%%%%%%%%%%%%%%%%%%%%%%%%%%%%%%%%%%%%%%%%%%%%%%%%%%%%%%%%%%%%%%%%%%%%%%%%

% References are important to the reader; therefore, each citation must be complete and correct. If at all possible, references should be commonly available publications.

\bibliographystyle{IEEEtran}
\bibliography{IEEEabrv, b_mybibfile}

\end{document}